\RequirePackage{fix-cm}
\documentclass[smallcondensed]{svjour3}     % onecolumn (ditto)
\smartqed  % flush right qed marks, e.g. at end of proof
\usepackage{graphicx}
\usepackage{epstopdf}
\usepackage{booktabs}
\usepackage{natbib}
\usepackage{amsmath}
\usepackage[linesnumbered,ruled,vlined]{algorithm2e}
\usepackage{amsfonts}       % blackboard math symbols

\SetKwInput{KwInput}{Input}                % Set the Input
\SetKwInput{KwOutput}{Output}              % set the Output

\begin{document}

\title{CPAS: the UK's National Machine Learning-based Hospital Capacity Planning System for COVID-19}
% \subtitle{Do you have a subtitle?\\ If so, write it here}

\titlerunning{CPAS: the UK's National ML-based Hospital Planning System for COVID-19}        % if too long for running head

\author{Zhaozhi Qian \and Ahmed M. Alaa \and Mihaela van der Schaar}


\institute{Zhaozhi Qian \at
           University of Cambridge \\
           \email{zhaozhi.qian@maths.cam.ac.uk}           
           \and
           Ahmed M. Alaa \at
           University of California, Los Angeles \\
           \email{ahmedmalaa@ucla.edu}
           \and
           Mihaela van der Schaar \at
           University of Cambridge \\
           University of California, Los Angeles \\
           \email{mv472@cam.ac.uk}
}

\date{Received: date / Accepted: date}
\maketitle

\begin{abstract}
The coronavirus disease 2019 (COVID-19) global pandemic poses the threat of overwhelming healthcare systems with unprecedented demands for intensive care resources. Managing these demands cannot be effectively conducted~without a nationwide collective effort that relies on data to forecast hospital demands on the national, regional, hospital and individual levels. To this end, we developed the {\it COVID-19 Capacity Planning and Analysis System} (CPAS) --- a machine learning-based system for hospital resource planning that we have successfully deployed at individual hospitals and across regions in the UK in coordination with NHS Digital. In this paper, we discuss the main challenges of deploying a machine learning-based decision support system at national scale, and explain how CPAS addresses these challenges by (1) defining the appropriate learning problem,~(2) combining bottom-up and top-down analytical approaches, (3) using state-of-the-art machine learning algorithms, (4) integrating heterogeneous data sources, and (5) presenting the result with an interactive and transparent interface. CPAS is one of the first machine learning-based systems to be deployed in hospitals on a national scale to address the COVID-19 pandemic --- we conclude the paper with a summary of the lessons learned from this experience.\\

\end{abstract}
\newpage
\section{Introduction}
\label{intro}
The coronavirus disease 2019 (COVID-19) pandemic poses immense challenges to healthcare systems across the globe --- a major issue faced by both policy~makers and front-line clinicians is the planning and allocation of scarce medical resources such as Intensive Care Unit (ICU) beds \citep{bedford2020covid}. In order to manage the unprecedented ICU demands caused by the pandemic, we need nationwide collective efforts that hinge on data to forecast hospital demands across various levels of regional resolution. To this end,~we~developed~the {\it COVID-19 Capacity Planning and Analysis System} (CPAS),~a~machine~learning-based tool that~has~been deployed to hospitals across the UK to assist the planning of ICU beds, equipment and staff \citep{CPAS}. CPAS is designed to provide actionable insights into the multifaceted problem of ICU capacity planning for various groups of stakeholders; it fulfills this goal by issuing accurate forecast for ICU demand over various~time horizons and resolutions. It makes use of the state-of-the-art machine learning techniques to draw inference from a diverse repository of heterogeneous data sources. CPAS presents its predictions and insights via an intuitive and interactive interface and allows the user to explore scenarios under different assumptions. \\

Critical care resources --- such as ICU beds, invasive mechanical ventilation and medical personnel --- are scarce, with much of the available resources being already occupied by severely-ill patients diagnosed with other diseases~\citep{ICU, ICM}. CPAS is meant to ensure a smooth operation of ICU by~anticipating~the required resources at multiple locations beforehand, enabling a timely management of these resources. While capacity planing has the greatest value at the peak of the pandemic, it also has important utility even \textit{after} the peak because it can help the hospitals to manage the transition from the COVID-19 emergency back to the normal business. During the pandemic, the vast majority of the healthcare resources were devoted to treating COVID-19 patients, so the capacity for treating other diseases were much reduced. It is therefore necessary to review and re-assign the resources after the COVID-19 trend starts to decline. CPAS is one of the first machine learning decision support systems deployed nationwide to manage ICU resources at different stages of the pandemic.\\ 

\begin{figure}
\centering
  \includegraphics[width=\columnwidth]{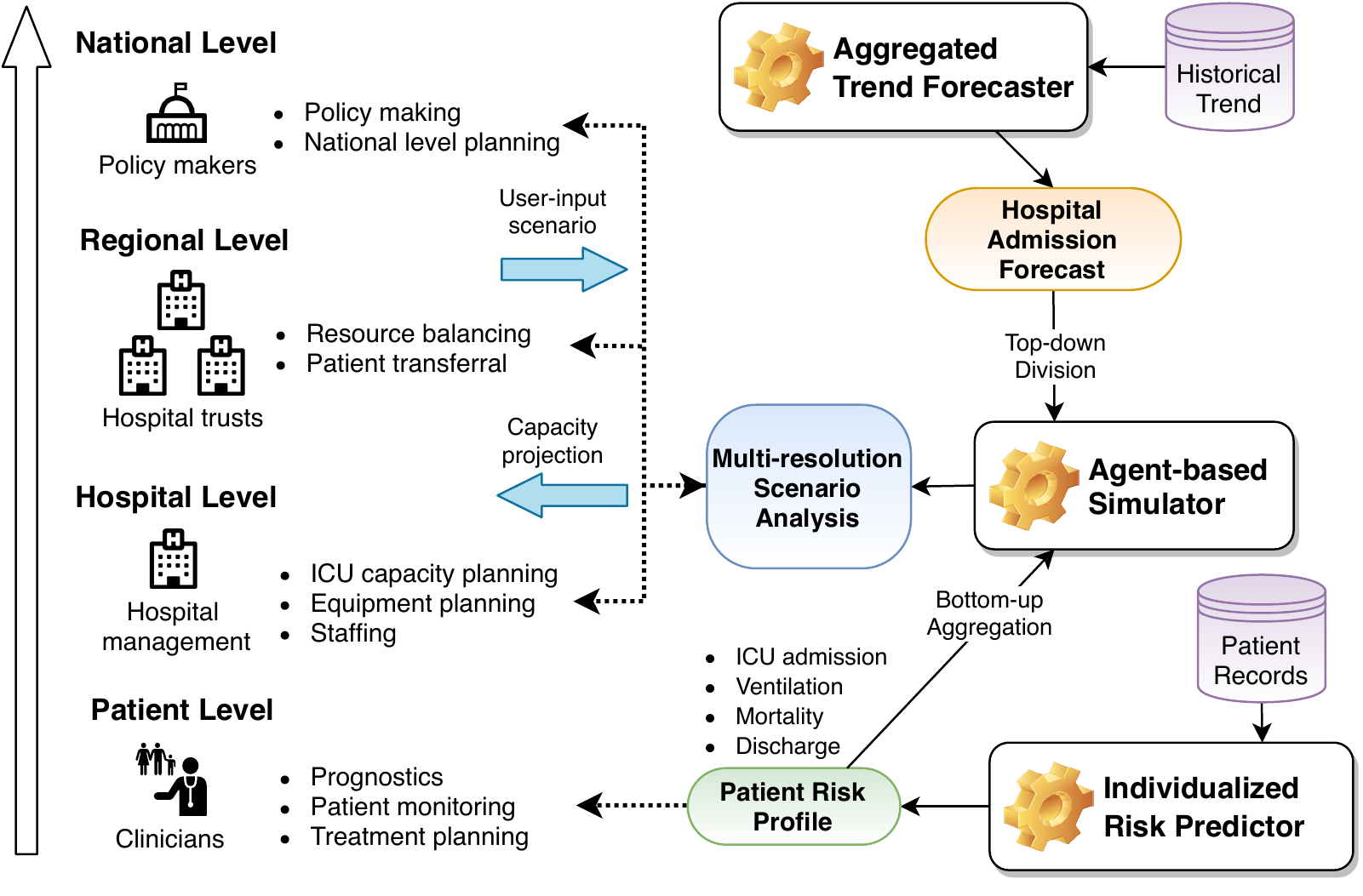}
\caption{
Illustration of how the different components in CPAS address the diverse needs of stakeholders on various levels. On the regional level, ``hospital trusts'' refers to the NHS foundation trusts, organizations that manage several hospitals in a region.
}
\label{fig:1}       % Give a unique label
\end{figure} 

% Although machine learning has long hold the promise to fundamentally change healthcare, there exists numerous practical challenges to implement and deploy a machine learning system \textit{rapidly at scale}. 
% We recognize five leading factors contributing to CPAS's fast and wide adoption. 
% Deploying machine learning system to large scale complex applications involves practical challenges often neglected by the academic literature. 
% In this section, we lay out the challenges we encountered in the development CPAS and gives an overview of the solution. 
CPAS is designed to serve the need of various groups of~stakeholders~involved in capacity planning at different \textit{levels} of geographical and administrative resolution. More specifically, as illustrated in Figure \ref{fig:1}, CPAS models ICU demands~on~the~(a) patient, (b) hospital, (c) regional and (d) national levels. On the~highest~level,~the system helps the {\it policy-makers} to make informed decisions by forecasting~the national trends for ICU demand under different scenarios. Secondly, \textit{regional} healthcare leadership can use CPAS as a ``load balancing'' tool as some hospitals may experience higher demand than others. Transferal of patients and resources can then be arranged accordingly. Next, hospital managers can use CPAS to plan ahead the \textit{local} ICU space, equipment and staff. Lastly, the front-line clinicians can make use of the tool to understand the risk profile of \textit{individual} patients. Importantly, these four groups of stakeholders require insights on different time horizons and levels of aggregation. CPAS addresses this challenge by combining the top-down projections from an ``aggregated trend forecaster'' and the bottom-up predictions from an ``individualized risk predictor''. Using agent-based simulation, CPAS is able to provide muti-resolution scenario analysis for individual, local, regional and national ICU demand.\\

As illustrated in Figure \ref{fig:1}, CPAS comprises an aggregated trend forecaster (top right) that issues overall projections of hospital admission trend. The projections are made on the hospital level over a time horizon of thirty days, and they are further aggregated to form the regional and national forecast.
CPAS also makes use of an individualized risk predictor (bottom right) that contains a suite of machine learning pipelines predicting the patient-level risk profiles of ICU admission, ventilator usage, mortality and discharge in the thirty-day period after hospital admission. The patient level risk profiles can be directly used by the clinicians to monitor patient status and make treatment plan. They can also be combined with the hospital admission forecast in the agent-based simulator (middle right) to perform scenario analysis for ICU capacity.
CPAS casts the complex practical problem of ICU capacity planning into a set of sub-problems, each of which can be addressed by machine learning. This divide-and-conquer approach makes CPAS a transparent solution rather than a monolithic black-box. The machine learning models underlying CPAS are trained using data that reflects the entire patient journey: from hospital admission, to ICU admission, to ventilation treatment, and finally to discharge or death. We will discuss the problem formulation in detail in section \ref{subsec:formulation}.\\

\noindent
{\bf The CPAS Machine Learning Model}\\

ICU planning for COVID-19 is a novel problem with few examples to learn from.
The spread of COVID-19 is modulated by the intrinsic characteristics of the novel virus as well as the unprecedented intervention policies by the government.  The need for deploying the system rapidly also means we cannot wait longer to collect more observations. As a result, the aggregated trend forecaster (Figure \ref{fig:1} top right) needs to learn the disease's transmission and progression characteristics from very limited data. To address this issue, we use a compartmental epidemiological model as a strong domain-specific \textit{prior} to drive the trend forecast. We integrated the prior in the proven framework of Bayesian hierarchical modelling and Gaussian processes to model the complex disease dynamics from few observations \citep{rasmussen2003gaussian}. We explain the aggregated trend forecaster in detail in section \ref{subsec:trend}.\\

% ICU planning for a novel pandemic also involves challenges for machine learning methods.
The individualized risk predictor (Figure \ref{fig:1} bottom right) is a production-level machine learning pipeline tasked to predict four distinct outcomes. 
For each outcome, it needs to address all stages of predictive modelling: missing data imputation, feature processing, prediction, and calibration.
For every stage, there are many machine learning algorithms to choose from, and for each algorithm there are multiple hyperparameters to tune.
However, using naive approach such as grid search to select algorithms and hyperparameters is very time-consuming, and it would hamper the rapid deployment of the system \citep{kotthoff2017auto}.
Since the pipeline configuration significantly affects the system's overall performance \citep{hutter2019automated}, we use a state-of-the-art AutoML tool designed for medical applications, AutoPrognosis \citep{alaa2018autoprognosis}, to address the algorithm and hyperparameter selection challenge for the individualized risk predictor (see section \ref{subsec:Individualized}).\\

It is inherently hard for a data-driven forecasting tool to accurately factor in the impact of unseen events (e.g. new social distancing policies). 
% Moreover, the distribution of patient feature shifts over time as the spread of pandemic progresses \citep{qian2020between}, which creates additional non-stationarity in the forecast problem.
CPAS uses the agent-based simulator (Figure \ref{fig:1} middle right) to allow the users to  explore scenarios under different assumptions about the policy impact.
The simulator first constructs a patient cohort that matches the feature distribution in the region of interest. It then uses the aggregated trend forecaster to determine the number of patients admitted to the hospital on each day based on the user's assumption about policy impact.
As is standard in agent-based simulation \citep{railsback2006agent}, each patient's outcomes are then simulated based on the risk profiles given by the individualized risk predictor. Finally, the simulated outcomes are aggregated to the desired level to form the scenario analysis. We introduce the details of the agent-based simulator in section \ref{subsec:sim}\\

\noindent
{\bf Challenges Associated with Building Practical Machine Learning-based Decision Support Systems}\\

Data integration poses a practical challenge for implementing a large-scale machine learning system.
As illustrated in Figure \ref{fig:1}, CPAS uses both patient records and the aggregated trends.
Typically, the patient level information is collected and stored by various hospital trusts in isolated databases with inconsistent storage formats and information schema.
It is a labour-intensive and time-consuming task to link and harmonize these data sources.
Furthermore, historical data that contains valuable information about the patients' pre-existing morbidities and medications, are archived on separated databases.
Accessing, processing and linking such historical data also proves to be challenging.
CPAS is trained using a data set constructed from four data sources. By breaking the data silos and linking the data, we draw diverse information covering the full spectrum of the patient health condition, which leads to informed and accurate prediction. 
We introduce the details of the datasets in section \ref{subsec:data}.\\

Presenting the insights in a user-friendly way is often neglected but vital to a machine-learning tool's successful adoption.
Ideally, the user interface should not only present the final conclusion but also the intermediate steps to reach the conclusion. The transparency of the system's internal working makes the system more trustworthy.
CPAS contains well-designed dashboards to display the outputs of the aggregated trend forecaster and the individualized risk predictors.
Thanks to the agent-based simulator, CPAS also allows the users to interactively explore the future scenarios by changing the underlying assumptions rather than presenting the forecast as the only possible truth.
We present an illustrative use case in section \ref{sec:usecase}.\\

% Thirdly, the usage of state-of-the-art algorithms including automated machine learning (AutoML), hierarchical Gaussian processes, and agent-based simulation enables CPAS to produce accurate data-driven scenario analysis.

% Last but not least, CPAS has a unique simulator interface that empowers the user to explore different scenarios autonomously rather than being told what to do. 
% It is not only a way to gain trust but also a channel to receive feedback.

% \section{ICU capacity planning: a challenging but rewarding problem}
% \label{sec:1}

% \subsection{Objectives and Requirements of CPAS}

The rest of the article is organized as follows: 
% In section \ref{sec:3} we formalize the ICU planning problem into a set of learning tasks and explain how CPAS solves them. 
After formulating the ICU planning problem into a set of learning tasks in section \ref{subsec:formulation}, we introduce the individualized risk predictor in section \ref{subsec:Individualized}, the aggregated trend forecaster in section \ref{subsec:trend}, and the agent-based simulator in section \ref{subsec:sim}.
We proceed to describe the data sources used in CPAS and the training procedure in section \ref{sec:4}.
After that, we present the offline evaluation results and discuss the need for online performance monitoring in section \ref{sec:exp}.
In section \ref{sec:usecase}, we demonstrate how CPAS works in action by going through an illustrative use case.
We conclude with the lessons learned in section \ref{sec:6}.

\section{CPAS: a system for ICU capacity planning}
\label{sec:3}

\subsection{Problem formulation}
\label{subsec:formulation}
To formulate the ICU capacity planning problem, we start by modelling the patients' arrival at each hospitals. Let $A_h(t)  \in \mathbb{N}$ be the number of  COVID-19 patients admitted to~a~given hospital $h \in \{1, \ldots, N\}$ on the $t^{th}$ day since the beginning of the outbreak.
Since not all hospitalized patients will require ICU treatment, we need to model the patient-level ICU admission risk to translate hospital admissions into ICU demand.
For a patient $i$ with a $D$-dimensional feature vector $X_i \in \mathbb{R}^D$, denote the event of ICU admission on the $\tau^{th}$ days after hospital admission as $Y_i(\tau) \in \{0, 1\}$.
Note that due to \textit{censoring}, we only observe $Y_i(\tau)$ for $\tau \in (0, \tau_i^*]$, where $\tau_i^*$ is the censoring time for patient $i$.
The ICU admission event $Y_i(\tau)$ directly translates into the ICU demand. To see this, consider a cohort of $A_h(t)$ patients admitted to hospital $h$ at time $t$. On a future date $t' > t$, the new ICU admission contributed by this cohort is given as:
\begin{equation}
    \text{ICU-inflow}(t \rightarrow t') = \sum_{i=1}^{A_h(t)} Y_i(t' - t),
\end{equation}
where we explicitly use $t \rightarrow t'$ to emphasize the date of hospital admission $t$ and the date of ICU demand $t'$. We can obtain the total ICU inflow on day $t'$ by summing over all historical patient cohorts with different hospital admission dates:
\begin{equation}
\label{eq:inflow}
    \text{ICU-inflow}(t') = \sum_{t<t'} \text{ICU-inflow}(t \rightarrow t') = \sum_{t<t'} \sum_{i=1}^{A_h(t)} Y_i(t' - t).
\end{equation}
It is apparent from the equation above that the ICU demand depends on two quantities: (1) the patient ICU risk profile $Y_i(\tau)$ and (2) the number of hospital admission $A_h(t)$ in the range of summation. Therefore, CPAS uses the \textbf{individualized risk predictor} to model $Y_i(\tau)$ and the \textbf{aggregated trend forecaster} to model $A_h(t)$ .
In addition to the ICU admission event, CPAS also contain models for three other clinical events: ventilator usage, mortality and discharge. We use $Y_i(\tau)$ to conceptually represent any outcome of interest when the context is clear. 

To build the CPAS individualized risk predictor, we consider a \textit{patient-level} data set $\mathcal{D}^P_{N, t}$ consisting patients from $N$ hospitals over a period of $t$ days, i.e., 
\begin{align}
\mathcal{D}^P_{N, t} := \left\{X_i, Y_i[1:\tau_i^*]  \right\}^{N_p}_{i=1},
\label{equation3}
\end{align}
where we use the square brackets to denote a sequence over a period of time i.e. $Y_i[1:\tau_i^*] := \left\{Y_i(1), \ldots, Y_i(\tau_i^*)\right\}$, and $N_p := \sum_{h=1}^N\sum_{j=1}^t A_h(j)$. 
CPAS learns the \textit{hazard function} for any patient with feature $X$ over a time horizon of $\tau \in (0, H]$:
\begin{align}
\widehat{h}(\tau) = \mathbb{P}(Y(\tau) = 1\ |\ X,\ \mathcal{D}^P_{N, t},\ Y(\tau') = 0, \forall \tau' < \tau)
\label{equation2}
\end{align}
Conceptually, the hazard function $\widehat{h}(\tau)$ represents the likelihood of ICU admission on day $\tau$ given the patient has not been admitted to the ICU in the first $\tau - 1$ days.
Therefore, for any patient who is already admitted to the hospital and registered in the system, CPAS is able to issue individual-level risk prediction based on the observed patient features $X_i$. Those individualized predictions can directly help the clinicians to monitor the patient status and design personalized treatment plan. 
The predicted risk profiles can also be aggregated to the hospital level to measure the ICU demand driven by the \textit{existing} patients.

To build the CPAS aggregated trend forecaster, we consider a \textit{hospital-level} dataset $\mathcal{D}^A_{N, t}$ for $N$ hospitals covering a period of $t$ days, i.e., 
\begin{align}
\mathcal{D}^A_{N, t} := \left\{A_h[1:t], M_h[1:t]\right\}^N_{h=1},
\label{equation3.1}
\end{align}
where the square brackets denote a sequence over time as before.
The dataset contains the number of hospital admissions $A_h(t)$, which we have defined previously, and the community mobility $M_h[1:t] := \{m_h(1) \ldots m_h(t)\}$ in the catchment of hospital $h$. 
The $m_h(t) \in \mathbb{R}^K$ is a $K$-dimensional real vector, with each dimension $k=1\ldots K$ reflecting the relative decrease of mobility in one category of places (e.g. workplaces, parks, etc) due to the COVID-19 containment and social distancing measures.
We used data for $N=94$ hospitals, each with $K=6$ categories of places. The details of the community mobility dataset is described in section \ref{subsec:mobility}.
For each hospital $h$, CPAS probabilistically forecasts the~trajectory~of~the~number~of COVID-19 admission within a future time horizon of $T$ days with a given community mobility, i.e., 
\begin{align}
\widehat{A}_h[t:t+T] = \mathbb{P}\Big[\,A_h[t:t+T]\,\,\Big|\,\,\underbrace{m_h(t),m_h(t+1),\ldots,\, m_h(t+T)}_{\mbox{\small \bf Future mobility $M_h[t:t+T]$}},\,\mathcal{D}^A_{N, t}\,\Big].
\label{equation4}
\end{align} 
The future mobility $M_h[t:t+T]$ depends on the intervention policies to be implemented in the future (e.g. schools to be closed in a week's time) and therefore may not be learnable from the historical data. For this reason, CPAS only models the \textit{conditional} distribution as in Equation \ref{equation4} and it allows the users to supply their own forecast of $M_h[t:t+T]$ based on their knowledge and expectation of the future policies. By supplying different values of $M_h[t:t+T]$, CPAS is able to project the hospital admission trend under these different scenarios. By default, CPAS extrapolates the community mobility using the average value of the last seven days i.e. $M_h(j) = \text{Avg}(M_h[t-8:t-1])$ for $j \in [t,\ t+T]$. In the following three sub-sections, we will introduce the individualized risk predictor, the aggregated trend forecaster, and the agent-based simulator. When multiple modelling approaches are possible, we will discuss their pros and cons and the reason why we choose a particular one in CPAS.

\begin{table}
\caption{The algorithms considered in each stage of the pipeline, which includes MICE \citep{buuren2010mice}, MissForest \citep{stekhoven2012missforest}, GAIN \citep{yoon2018gain}, PCA, Fast ICA \citep{hyvarinen1999fast}, Recursive elimination \citep{guyon2002gene}, Elastic net \citep{zou2005regularization}, Random forest \citep{liaw2002classification}, Xgboost \citep{chen2016xgboost}, Muti-layer Perceptron (MLP) \citep{hinton1990connectionist}, Isotonic regression \citep{de1977correctness}, Bootstrap  \citep{chernick2011bootstrap}, Platt scaling \citep{platt1999probabilistic}. }
\label{tab:1}       % Give a unique label
\begin{tabular}{@{}llll@{}}
\toprule
\textbf{Imputation} & \textbf{Feature Selection} & \textbf{Prediction} & \textbf{Calibration} \\ \midrule
Median              & No selection               & Elastic net         & Isotonic regression  \\
MICE               & PCA                        & Random forest       & Bootstrap             \\
MissForest           & Fast ICA                   & Xgboost             & Platt scaling    \\
GAIN                & Recursive elimination      & MLP                 &                      \\ \bottomrule
\end{tabular}
\end{table}

\subsection{Individualized risk prediction using automated machine learning}
\label{subsec:Individualized}

In order to learn the hazard function defined over a period of $\tau \in (0, H]$ days (Equation \ref{equation2}),
the individualized risk predictor contains $H$ \textit{calibrated binary classification pipelines}, trained independently and each focusing on a single time step, i.e. $\widehat{h}(\tau) = P_\theta^\tau(X)$ for $\forall \tau \in (0, H]$. 
We will discuss the concept of \textit{pipelines} later in this section. For now, the readers can assume $P_\theta^\tau(X)$ to be a binary classifier with hyperparameters $\theta$.
By training separate pipelines for each time step, we do not assume that the hazard function follows any specific functional form, which adds to the flexibility to model complex disease progression.
The training data for each time step $\tau$, denoted as $\mathcal{D}^P_{N, t}(\tau)$ is derived from the full patient-level dataset $\mathcal{D}^P_{N, t}$ as follows:
\begin{align}
\mathcal{D}^P_{N, t}(\tau)= \{ (X_i, Y_i(\tau)) \ |\ (X_i, Y_i[1:\tau_i^*]) \in \mathcal{D}^P_{N, t}, \ \tau_i^* \ge \tau ,\ Y_i(\tau') = 0, \forall \tau' < \tau\}
\label{eq:data}
\end{align}
The first condition $(X_i, Y_i[1:\tau_i^*]) \in \mathcal{D}^P_{N, t}$ simply states that the patient $i$ is in the full dataset. The second condition $\tau_i^* \ge \tau$ ensures that the $Y_i(\tau)$ is observed. In other words, the status of patient $i$ at time $\tau$ is not \textit{censored}. The last condition $Y_i(\tau') = 0, \forall \tau' < \tau$ arises from the definition of the hazard function in Equation \ref{equation2}. 
Jointly these three conditions ensure that the binary classifier trained on $\mathcal{D}^P_{N, t}(\tau)$ predicts the hazard function $\widehat{h}(\tau)$ at time $\tau$.

\begin{figure}
  \includegraphics[width=1\columnwidth]{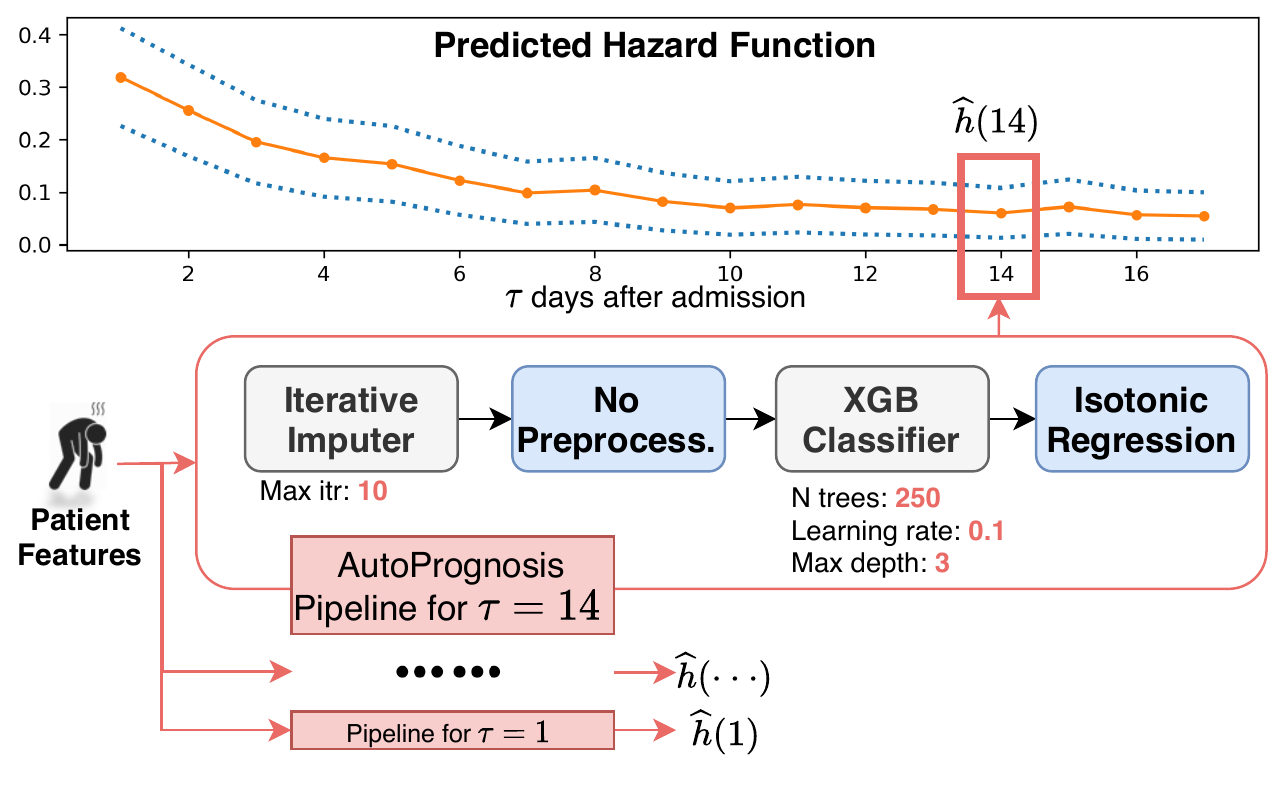}
\caption{Schematic depiction for the individualized risk predictor. A patient's feature is fed into multiple pipelines in parallel. Each pipeline estimates the hazard function $\widehat{h}(\tau)$ at a different time step $\tau \in (0, H]$.
The pipeline configuration specifies the algorithms and the associated hyperparameters. The configurations are determined by AutoPrognosis using equation \ref{eq:cv} and may vary across $\tau$.}
\label{fig:individual}       % Give a unique label
\end{figure}

A machine learning \textit{pipeline} consists of multiple stages of predictive modelling.
Let $\mathcal{F}_d, \mathcal{F}_f ,\mathcal{F}_p,\mathcal{F}_c$ be the sets of all missing data imputation, feature processing, prediction, and calibration algorithms
we consider (Table \ref{tab:1}) respectively.
A pipeline $P$ is a tuple of the form:
\begin{align}
P = (F_d, F_f ,F_p,F_c),
\label{eq:pipeline}
\end{align}
where $F_i \in \mathcal{F}_i$, $\forall i \in \{d, f, p, c\}$.
The space of all possible pipelines is given by $\mathcal{P} = \mathcal{F}_d \times \mathcal{F}_f \times \mathcal{F}_p \times \mathcal{F}_c$. Thus, a pipeline consists of a selection of algorithms from each column of Table \ref{tab:1}. For example, 
$P$ = (MICE, PCA, Random Forest, Platt scaling). The total number of pipelines we consider is $|\mathcal{P}| = 192$.
While $P$ specifies the ``skeletion'' of the pipeline, we also need to decide the hyperparameter configuration of its constituent algorithms.
Let $\Theta = \Theta_d \times \Theta_f \times \Theta_p \times \Theta_c$ be the space of all hyperparameter configurations. Here $\Theta_v = \bigcup_a \Theta^a_v$ for $v \in  \{d, f, p, c\}$ with $\Theta^a_v$ being the space of hyperparameters
associated with the $a^{th}$ algorithm in $\mathcal{F}_v$.
Therefore, a fully specified pipeline configuration $P_\theta \in \mathcal{P}_\Theta$ determines the selection of algorithms $P \in \mathcal{P}$ and their corresponding hyperparameters $\theta \in \Theta$.

\textbf{Why not use survival analysis?} Some readers might have noticed that our problem formalism is similar to \textit{survival analysis}, which also deals with time-varying outcome (survival) and considers censored data. When developing CPAS, we have considered this class of models. In fact, we have used the Cox proportional hazard model as a baseline benchmark (see section \ref{sec:exp}). However, two factors discouraged us from further pursuing survival models. Firstly, the available implementations of survival models are not as abundant as classifiers and they are often immature for industrial scale applications. Secondly, many modern machine-learning powered survival models do not make the proportional hazards assumption \citep{polsterl2016efficient,van2011learning,hothorn2006survival}. The expense of relaxing assumption is that these models are often not able to estimate the full survival function and measure the absolute risk at a given time. There are recent works in survival analysis trying to address this issue \citep{lee2019temporal}, but it is still an open research area.

\textbf{Training the individualized risk predictor}. In training, we need to find the best pipeline configuration $P_{\theta^*}^* \in \mathcal{P}_\Theta$ that empirically minimizes the $J$-fold cross-validation loss:
\begin{align}
P_{\theta^*}^* = \text{argmin}_{P_{\theta} \in \mathcal{P}_\Theta} \sum_{j=1}^J \mathcal{L}\big(P_{\theta},\ \mathcal{D}^P_{N, t}[{\text{Train}(j)}],\ \mathcal{D}^P_{N, t}[{\text{Val}(j)}]\big),
\label{eq:cv}
\end{align}
where $\mathcal{L}$ is the loss function (e.g. Brier score), and $\mathcal{D}^P_{N, t}[{\text{Train}(j)}]$ and $\mathcal{D}^P_{N, t}[{\text{Val}(j)}]$ are the training and validation splits of the patient-level dataset $\mathcal{D}^P_{N, t}$.
Note that this is a very hard optimization problem due to three facts (1) the space of all pipeline configurations $\mathcal{P}_\Theta$ is very high dimensional, (2) the pipeline stages interact with each other and makes the problem not easily decomposable, and (3) evaluating the loss function via cross validation is a time-consuming operation.

Instead of relying on heuristics, we apply the state-of-the-art automated machine learning tool AutoPrognosis  to  configure all stages of the pipeline jointly \citep{alaa2018autoprognosis}.
AutoPrognosis  is based on Bayesian optimization (BO), an optimization framework that has achieved remarkable success in optimizing black-box functions with costly evaluations as compared to simpler approaches such as grid and random search \citep{snoek2012practical}.
The BO algorithm used by AutoPrognosis implements a sequential \textit{exploration-exploitation} scheme in which balance is achieved between exploring the predictive power of new pipelines and re-examining the utility of previously explored ones.
To deal with high-dimensionality, AutoPrognosis  models the “similarities” between the pipelines’ constituent algorithms via a sparse additive kernel with a Dirichlet prior.
When applied to related prediction tasks, AutoPrognosis  can also be warm started by calibrating the priors using an meta-learning algorithm that mimics the empirical Bayes method, further improving the speed.
For more technical details, we refer the readers to \citet{alaa2018autoprognosis}.

The AutoPrognosis framework has been successfully applied to building prognostic models for Cystic Fibrosis and Cardiovascular Disease \citep{alaa2018prognostication, alaa2019cardiovascular}. Implemented as a Python module, it supports 7 imputation algorithms, 14 feature processing algorithms, 20 classification algorithms, and 3 calibration methods; a design space which corresponds to a total of 5,880 pipelines. 
We selected a subset of most commonly used and well-understood algorithms for CPAS (Table \ref{tab:1}), all of which have achieved numerous success in machine learning applications.

\subsection{Trend forecast using hierarchical Gaussian process with compartmental prior}
\label{subsec:trend}

CPAS uses a hierarchical Gaussian process with compartmental prior (HGPCP) to forecast the trend of hospital admission. 
HGPCP is a Bayesian model that combines the data-driven Gaussian processes (GP) and the domain-specific compartmental models \citep{li1995global, rasmussen2003gaussian}. 

\begin{figure}
  \includegraphics[width=1\columnwidth]{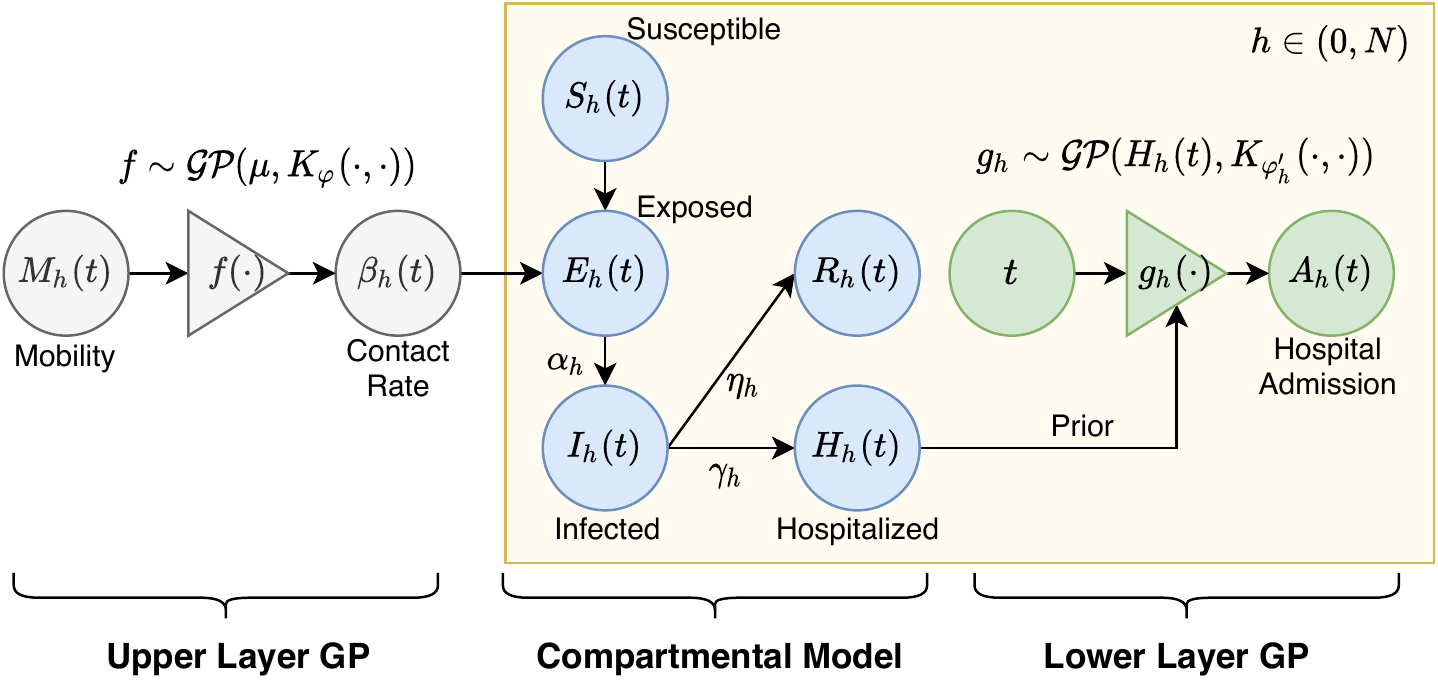}
\caption{Pictorial illustration of HGPCP. Left to right: The upper-layer GP $f(\cdot)$ models the contact rate $\beta_h(t)$ based on community mobility $M_h(t)$.
The compartmental model gives the deterministic trajectory of the five compartments based on $\beta_h(t)$.
The lower layer GP uses the hospitalized compartment $H_h(t)$ as prior and predicts the hospital admission.
}
\label{fig:3}       % Give a unique label
\end{figure}

The compartmental model is a family of time-honoured mathematical models designed by domain experts to model epidemics \citep{kermack1927contribution}. 
We use a specific version to model hospital admission \citep{hethcote2000mathematics, osemwinyen2015mathematical}.
As illustrated in Figure \ref{fig:3}, the compartmental model partitions the whole population containing $C_h$ individuals into disjoint \textit{compartments}: Susceptible $S_h(t)$, Exposed $E_h(t)$, Infectious $I_h(t)$, Hospitalized $H_h(t)$, and Recovered or died outside hospital $R_h(t)$, for $\forall h \le N$.
At any moment in time, the sum of all compartments is equal to the size of population i.e. $S_h(t) + E_h(t) + I_h(t) + H_h(t) + R_h(t) = C_h$ for $\forall t > 0$.
As the pandemic unfolds, the sizes of the compartments change according to the following differential equations:
\begin{equation}
\begin{split}
\frac{dS_h}{dt} = -\beta_h(t) S_h I_h&,\,\,\, \frac{dE_h}{dt} = \beta_h(t) S_h I_h - \alpha_h E_h,\,\,\, \frac{dI_h}{dt} = \alpha_h E_h - \gamma_h I_h, \\
& \frac{dH_h}{dt} = \eta_h \gamma_h I_h,\,\,\, \frac{dR_h}{dt} = (1 - \eta_h) \gamma_h I_h,
\label{eq:compartments}
\end{split}
\end{equation}
where $\alpha_h$, $\gamma_h$, $\eta_h$ are hospital-specific parameters describing various aspects of the pandemic, and $\beta_h(t)$ is the \textit{contact rate} parameter that varies across hospitals and time.
Among all the parameters, the contact rate $\beta_h(t)$ is of special importance for two reasons: (1) $\beta_h(t)$ is the coefficient of the only non-linear term in the equation: $S_h \cdot I_h$. Therefore the nonlinear dynamics of the system heavily depends on $\beta_h(t)$. (2) the contact rate changes over time according to how much people travel and communicate. It is therefore heavily influenced by the various intervention policies.

To model the time-varying contact rate, the \textbf{upper layer} of HGPCP utilizes a function $f$ drawn from a GP prior with mean $\mu$ and covariance kernel $K_\varphi(\cdot, \cdot)$. The function $f$ maps community mobility $M_h(t)$ to the time-varying contact rate $\beta_h(t)$ as follows:
\begin{equation}
 f \sim \mathcal{G}\mathcal{P}(\mu, K_\varphi(\cdot, \cdot)), \,\;\;
 \beta_h(t) = f(M_h(t)), 
\label{eq:upper}
\end{equation}
where $\mu$ and $\varphi$ are hyperparameters of the GP. It is worth noticing that the upper layer of HGPCP is shared across hospitals.
With the contact rate given by Equation \ref{eq:upper}, the differential equations (\ref{eq:compartments}) can be solved to obtain the trajectory of all compartments. We used Euler's method \citep{EULER} to solve these equations but other solvers are possible.
The \textbf{lower layer} of HGPCP consists of $N$ independent GPs that use the hospitalized compartment $H_h(t)$ as a \textit{prior} mean function and predicts hospital admissions over time:
\begin{equation}
 g_h \sim \mathcal{G}\mathcal{P}(H_h(t), K_{\varphi'_h}(\cdot, \cdot)),\;\;\;A_h(t) = g_h(t),\;\;\;\forall h \le N 
\label{eq:lower}
\end{equation}
where $\varphi'$ is the kernel hyperparameters. We use the radial basis function as the kernel for both upper and lower layers.
Combining equation \ref{eq:compartments}, \ref{eq:upper}, and \ref{eq:lower}, the prediction problem \ref{equation4} can be formulated in the following posterior predictive distribution:
\begin{align}
\widehat{A}_h[t:t+T] = \int \underbrace{\mathbb{P}\big(\,\mathcal{A}_h[t:t+T]\,\big|\,\mathcal{A}_h[1:t], H_h\big)}_{\mbox{\footnotesize {\bf Lower layer GP}}}\cdot 
\underbrace{\mathbb{P}\big(\,H_h\,\big|\,\beta_h, \theta_h \big)}_{\mbox{\footnotesize {\bf Compartmental model}}} \\
\underbrace{d\mathbb{P} \big(\,\beta_h\,\big|\,\mathcal{D}^A_{N,t}, \mathcal{M}_{h}[t:t+T]\big)}_{\mbox{\footnotesize {\bf Upper layer GP}}} \mathbb{P}(\theta_h), \nonumber 
\end{align}
where $H_h := \{H_h(1), \ldots, H_h(t+T) \}$, $\beta_h := \{\beta_h(1), \ldots, \beta_h(t+T) \}$, and $\theta_h = (\alpha_h, \gamma_h, \eta_h, \varphi'_h)$.
In this equation, the GP posterior terms can be computed analytically \citep{rasmussen2003gaussian} while the compartmental model term has no closed form solution due to its nonlinearity. Therefore, we evaluate the integral via Monte Carlo approximation and derive the mean and quantiles from the Monte Carlo samples. 

\textbf{What is the advantage over a standard GP?} Compared with a standard zero-mean GP, HGPCP is able to use the domain knowledge encoded in the compartmental models to inform the prediction.
Since the spread of a pandemic is a highly non-stationary process: after the initial phase of exponential growth, the trend flattens and gradually reaches saturation, 
extrapolating observed data without considering the dynamics of the pandemic spread is likely to be misleading.
The injection of domain knowledge is especially helpful at the early stage of the pandemic where little data are available. 

\textbf{What is the advantage over a compartmental model?} Compared with the compartmental models, HGPCP uses the GP posterior to make prediction in a data-driven way. Since HGPCP only uses the compartmental trajectories as a prior, it is less prone to model mis-specifications. 
Moreover, HGPCP is able to quantify prediction uncertainties in a principled way by computing the predictive posterior, whereas the compartmental models can only produce trajectories following deterministic equations. 
In capacity planning applications, the ability to quantify uncertainty is especially important.

\textbf{Training HGPCP.} So far, we have assumed that the hyperparameters $\alpha = (\varphi, \mu)$ of the upper layer GP is given. In practice, we optimize these hyperparameters by maximizing the log-likelihood function on the training data $\mathcal{D}^A_{N,t}$:
\begin{align}
\mathcal{L}(\mathcal{D}^A_{N, t}\,|\,\alpha) := \log \int \, \prod^N_{h=1} \, \mathbb{P}\big(\,\mathcal{A}_h[1:t]\,\big|\, H_h\big)\cdot \mathbb{P}\big(\,H_h\,\big|\,\mathcal{M}_h[1:t],\alpha\big)\,dH_h,
\label{eq:log-lik}
\end{align}
and $\alpha^* = \text{argmax}_\alpha \mathcal{L}(\mathcal{D}^A_{N, t}\,|\,\alpha)$.
Since the integral in (\ref{eq:log-lik}) is intractable,  we resort to a variational inference approach for optimizing the model's likelihood \citep{ranganath2014black,wingate2013automated}. That is, we maximize the evidence lower bound (ELBO) on (\ref{eq:log-lik}) given by:   
\begin{align}
\mbox{ELBO}_i(\alpha, \phi) = \mathbb{E}_{\mathbb{Q}} \left[\log \mathbb{P}\big(\,\mathcal{A}_h[1:t], H_h\,\big|\,\alpha\big) - \log \mathbb{Q}\big(\,H_h\,\big|\, \mathcal{A}_h[1:t], \phi\big) \right], \nonumber  
\label{equation10}
\end{align}
where $\mathbb{Q}(.)$ is the variational distribution parameterized by $\phi$ with conditioning on $\mathcal{M}_h[1:t]$ omitted for notational brevity.
We choose a Gaussian distribution for $\mathbb{Q}(.)$, which simplifies the evaluation the ELBO objective and its gradients. We use stochastic gradient descent via ADAM algorithm to optimize the ELBO objective \cite{ADAM}.

The trained HGPCP model can issue forecasts on hospital level. To get regional or national level forecast, denote the set of hospitals in the region $r$ as $\mathcal{H}_r$, and the regional forecast is obtained by taking summation of the constituent hospitals i.e. $\widehat{A}_r(t) = \sum_{h \in \mathcal{H}_r} \widehat{A}_h(t)$ for $\forall t \in (0, T]$.

\subsection{Agent-based simulation}
\label{subsec:sim}
\begin{algorithm}
\label{alg:1}
\DontPrintSemicolon
  \KwInput{The set of hospitals in the area of interest $\mathcal{H}_r$}
  \KwInput{Mobility trend $M(t)$ for $t \in (0, H]$}
  \KwOutput{ICU admission over time $\text{ICU-Inflow}(d)$, $d = 1, \ldots$}
  
  Initialize $\text{ICU-Inflow}(d) = 0$ for $d \in (0, H]$
  
  Calculate patient feature distribution $\hat{p}(X)$ 
  
  Daily hospital admissions $A(t) = \text{TrendForecaster}(\mathcal{H}_r, M(1), M(2), \ldots, M(t-1))$
  
  \For{$t \in (0, H]$} 
  {
    \For{$i \in (0, A(t)]$}
      {
        $X_{i} \sim \hat{p}(X)$  \tcp*{the i\textsuperscript{th} patient admitted on day $t$}
        
        \For{$\tau \in (0, T]$}
        {
          $p_{i}^\tau = P_\theta^\tau(X_{i})$ \tcp*{ICU risk on day $\tau$ after admission}
          
          $s_{i}^\tau \sim \text{Bernoulli}(p_{i}^\tau)$
          
          \If{$s_{i}^\tau = 1$}
          {
            $\text{ICU-Inflow}(t + \tau) \mathrel{{+}{=}} 1$
            
            Continue \tcp*{Skipping the rest of the $\tau$-loop}
          }
        }
      }
  }
\caption{Agent-based ICU inflow simulation}
\end{algorithm}

The individualized risk predictor can be used to predict the ICU demand caused by the patients who are currently staying in the hospital and whose features $X_i$ are known. 
However, the total ICU demand is also driven by the patients arriving at the hospital in the future whose features are currently unknown.
From discussions with the stakeholders, we understand that the ICU demands from future patients are especially important for setting up new hospital wards. 
CPAS uses agent-based simulation \citep{railsback2006agent} to perform scenario analysis for future patients. 

To estimate the ICU demand caused by future patients, we need to answer two questions: (1) how many new patients will be admitted to the hospital in the future? and (2) what will be the risk profile of these patients? The aggregated trend forecaster is precisely designed to answer the first question whereas the individualized risk predictor can answer the second question if the patient features were known. We can use the empirical feature distribution of the existing patients to approximate the distribution of the future patients. The empirical feature distribution is defined as $\hat{p}(X) = \sum_{j=1}^{N_P} \mathcal{I}(X_j = X) / N_P$, where $\mathcal{I}(\cdot)$ is the indicator function and $N_p$ is the number of patients in the region of interest.

The algorithm is detailed in Algorithm \ref{alg:1}.
The simulator takes two sets of users inputs.
The user first specifies the level of resolution (hospital, regional, or national) and chooses what hospital or region to examine from a drop down list. 
% Then the user can optionally specify the patient feature distribution $p(X)$ including age distribution, sex ratio and prevalence of various comorbidities. 
% By default, the feature distribution is assumed to follow the historical empirical distribution in the region, from which we can draw samples with replacement \citep{chernick2011bootstrap}.
% For user specified distributions, the age is modelled as a log-normal distribution because we find it fits well to the data. The user needs to specify the mean and standard deviation of the log-normal distribution.
% All other features are binary variables and we modelled them as independent Bernoulli variables with the user-specified means. Although the independence assumption is very strong, we did not find a better way to allow the users to specify the distributions in more details (e.g. to include the correlation between features) without overwhelming them.
Next, the user specifies the future community mobility trend $M(t)$ to reflect the government plan to maintain or easy social distancing. The default value of $M(t)$ is a constant value given by the average over the last seven days.

In the simulation the aggregated trend forecaster first generates a forecast of daily hospital admissions $A(t)$. 
It then generates a patient cohort with $A(t)$ patients arriving at the hospital on day $t$, whose features are independently sampled from distribution $\hat{p}(X)$.
Next, the individualized risk predictor $P_\theta^\tau$ obtains the calibrated ICU admission risk on the $\tau$-th day after hospital admission. 
Based on the risk scores, we take a Monte Carlo sample to decided when each patient will be admitted to ICU and update the total ICU inflow accordingly.
The above procedure can be repeated many times to obtain the Monte Carlo estimate of variation in ICU inflow. 
The ICU outflow due to discharge or death as well as ventilator usage can be derived in a similar fashion, and is omitted for brevity.

\section{Training and deploying CPAS}
\label{sec:4}
\subsection{Dateset}
\label{subsec:data}
\begin{figure}
  \includegraphics[width=1\columnwidth]{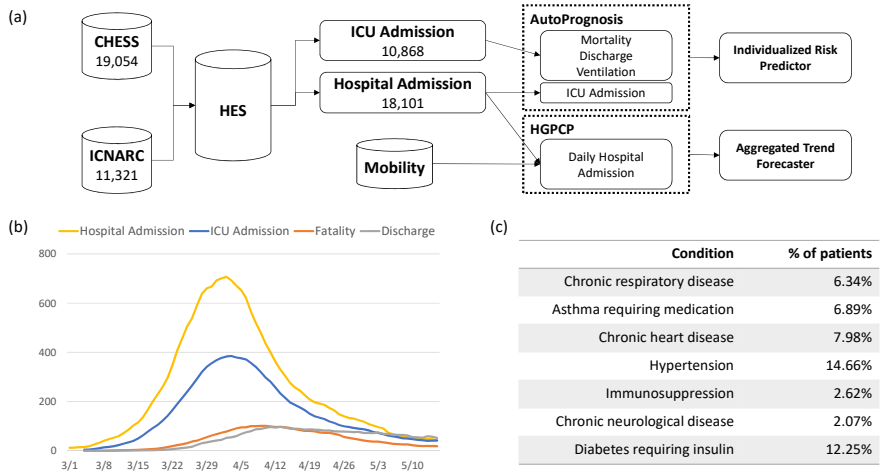}
\caption{Illustration of the CPAS datasets and the training set up. (a) CHESS and ICNARC data are joined and linked to HES to form the hospital patient data (18,101 cases) and the ICU patient data (10,868 cases).
AutoPrognosis uses these two patient level datasets to train the various predictive pipelines in the individualized risk predictor.
The aggregated hospital admission data together with the community mobility data empowers HGPCP to forecast the trend of admission.
(b) The daily hospital admission, ICU admission, fatalities and discharges as recorded in the CPAS data set. (c) The prevalence of comorbidities and complications of hospitalized COVID-19 patients. }
\label{fig:4}       % Give a unique label
\end{figure}

CPAS relies on three distinct sources of patient-level data, each covering a unique aspect of patient health condition (Figure \ref{fig:4}). 
CPAS also makes use of community mobility trend data to issue aggregated trend forecast.
The details of these data sources are described in the following sub-sections.
The summary statistics of the data sets as of May 20\textsuperscript{th} are shown in Figure \ref{fig:4}.

\subsubsection{COVID-19 hospitalizations in England surveillance system (CHESS)}
COVID-19 Hospitalizations in England Surveillance System (CHESS) is a surveillance scheme for monitoring hospitalized COVID-19 patients. The scheme has been created in response to the rapidly evolving COVID-19 outbreak and has been developed by Public Health England (PHE). It has been designed to monitor and estimate the impact of COVID-19 on the population in a timely fashion, to identify those who are most at risk and evaluate the effectiveness of countermeasures.

CPAS uses the de-identified CHESS data updated daily from 8\textsuperscript{th} February (data collection start), which records COVID-19 related hospital and ICU admissions from 94 NHS trusts across England. 
The data set features comprehensive information on patients' general health condition, COVID-19 specific risk factors (e.g. comorbidities), basic demographic information (age, sex, etc.), and tracks the entire patient treatment journey: when each patient was hospitalized, whether they were admitted to the ICU, what treatment (e.g. ventilation) they received, and their mortality or discharge outcome.

\subsubsection{Intensive care national audit \& research centre database (ICNARC)}

Intensive care national audit \& research centre (ICNARC) maintains a database on patients critically ill with confirmed COVID-19. The data are collected from ICUs participating in the ICNARC Case Mix Programme (covering all NHS adult, general intensive care and combined intensive care/high dependency units in England, Wales and Northern Ireland, plus some additional specialist and non-NHS critical care units).
CPAS uses the de-identified ICNARC data
which contains detailed measurements of ICU patients' physiological status (PaO2/FiO2 ratio, blood pH, vital signals, etc.) in the first 24 hours of ICU admission. It also records each patient's organ support status (respiratory support, etc). ICNARC therefore provides valuable information about the severity of patient condition. 

\subsubsection{Hospital episode statistics (HES)}

Hospital Episode Statistics (HES) is a database containing details of all admissions, A and E attendances and outpatient appointments at NHS hospitals in England.
We retrieve HES records for patients admitted to hospital due to COVID-19.
While the HES record contains a wide range of information about an individual patient, CPAS only make use of the clinical information about disease diagnosis.
All other information are discarded during the data linking process to maximally protect privacy.
HES is a valuable data source because it provides comprehensive and accurate information about patients' pre-existing medical conditions, which are known to influence COVID-19 mortality risk.

\subsubsection{Community mobility reports}
\label{subsec:mobility}
In addition to the above patient-level data, we also used the COVID-19 Community Mobility Reports produced by Google \citep{Mobility}. 
The dataset tracks the movement trends over time by geography, across six different categories of places including (1) retail and recreation, (2) groceries and pharmacies, (3) parks, (4) transit stations, (5) workplaces, and (6) residential areas, resulting in a $K=6$ dimensional time series of community mobility.
It reflects the change in people's behaviour in response to the social distancing policies.
We use this dataset to inform the prediction of contact rates over time.
The dataset is updated daily starting from the start of the pandemic.

% \begin{figure}
% % Use the relevant command to insert your figure file.
% % For example, with the graphicx package use
%   \includegraphics[width=1\columnwidth]{CPAS_train.pdf}
% % figure caption is below the figure
% \caption{An illustration of how each component of CPAS is trained.}
% \label{fig:5}       % Give a unique label
% \end{figure}

\subsection{Training procedure}

By linking the three patient-level data sources described in the last section, we create two  data sets containing hospitalized and ICU patients respectively (Figure \ref{fig:4}).
For the ICU patients, we use AutoPrognosis to train three sets of models for mortality prediction, discharge prediction and ventilation prediction over a maximum time horizon of 30 days. 
For hospitalized patients, we carry out a similar routine to train predictors for ICU admission.
All the AutoPrognosis pipelines are then deployed to form the individualized risk predictor in CPAS.
Furthermore, we aggregate the patient-level data to get the daily hospital admission on hospital level. 
The data set is then combined with the community mobility report data to train the aggregated trend forecaster.

To make sure CPAS uses the most up-to-date information, we retrain all models in CPAS daily. 
The re-training process is automatically triggered whenever a new daily batch of CHESS or ICNARC arrives.  
Data linking and pre-processing is carried out on a Spark cluster with 64 nodes, and usually takes less than an hour to complete. 
After that, model training is performed on a HPC cluster with 116 CPU cores and 348 GB memory, and typically finishes within three hours.
Finally, the trained models are deployed to the production server and the older model files are archived.

\section{Evaluation and performance monitoring}
\label{sec:exp}
\subsection{Offline evaluation}

\begin{table}
\caption{ Performance in forecasting individualized risk profile using different feature sets and algorithms measured by AUC-ROC. The results in \textbf{bold} are significantly better than the rest.}
\label{tab:2}       % Give a unique label
\begin{tabular}{@{}lllll@{}}
\toprule
Model          & Feature      & ICU Admission      & Mortality         & Ventilation       \\ \midrule
AutoPrognosis  & All features & \textbf{0.835 $\pm$ 0.001}  & \textbf{0.871 $\pm$ 0.002} & \textbf{0.771 $\pm$ 0.002} \\
AutoPrognosis  & CHESS only   & 0.781 $\pm$  0.002 & 0.836 $\pm$ 0.002 & 0.754 $\pm$ 0.003                \\
AutoPrognosis  & Demographics & 0.770 $\pm$  0.002 & 0.799 $\pm$ 0.003 & 0.702 $\pm$ 0.003               \\
Cox PH Model   & All features & 0.771 $\pm$  0.002 & 0.773 $\pm$ 0.003 & 0.690 $\pm$ 0.003 \\
Charlson Index & -            & 0.556 $\pm$  0.013 & 0.596 $\pm$ 0.002 & 0.530 $\pm$ 0.006               \\ \bottomrule
\end{tabular}
\end{table}

In the offline evaluation, we first validated two hypothesis about the individualized risk predictor: (1) using additional patient features improves risk prediction and (2) the AutoPrognosis pipeline significantly out performs the baseline algorithms. We performed 10-fold cross validation on the data available as of March 30 (with 1200 patients in total) and evaluated the AUC-ROC score on the $\tau=7$ day risk prediction. 
We compared AutoPrognosis to two benchmarks that are widely used in clinical research and Epidemiology: the Cox proportional hazard model \citep{cox1972regression} and the Charlson comorbidity index \citep{charlson1994validation}.
The results are shown in Table \ref{tab:2}. We can clearly see that there is a consistent gain the predictive performance when more features are included in the AutoPrognosis model and AutoPrognosis significantly outperforms the two benchmarks. 

Next, we turned to the aggregated trend forecaster and validated the following two hypothesis: (1) HGPCP outperforms the zero-mean GP due to a more sensible prior (the compartmental model), and (2) HGPCP outperforms the compartmental model because GP reduces the risk of model mis-specification.
We evaluate the accuracy of the 7-day projections issued at three stages of the pandemic: before the peak of infections (March 23), in the midst of the peak (March 30), and in the “plateauing” stage (April 23).
Accuracy was evaluated by computing the mean absolute error between true and predicted daily hospital admission throughout the forecasting horizon, i.e., 
$\sum_{t=1}^{7}|A_h(t) - \widehat{A}_h(t)| / 7$. 
In table \ref{tab:3} we report the performance for the five hospitals with most COVID-19 patients as well as the national level projection produced by aggregating all hospital level forecasts.
We observe that HGPCP consistently outperforms the benchmarks both on hospital and national level across different stages of the pandemic.

\begin{table}
\caption{ Performance in forecasting hospital admission. The candidate models are CPAS (HGPCP), GP (zero-mean GP) and CM (compartmental models). The first five rows refer to the performance in the five hospitals with most admitted patients. The last row refers to the national total admission. The lowest error for each task is bolded.}
\label{tab:3}       % Give a unique label
\begin{tabular}{@{}lccccccccc@{}}
\toprule
         & \multicolumn{3}{c}{Mar. 23 before peak} & \multicolumn{3}{c}{Mar. 30 at peak} & \multicolumn{3}{c}{Apr. 23 after peak} \\
         & CPAS      & GP      & CM         & CPAS     & GP     & CM        & CPAS      & GP      & CM         \\ \midrule
STH      & \textbf{3.01}       & 5.32         & 7.39       & \textbf{11.49}     & 13.79       & 14.72     & \textbf{3.18}       & 6.45         & 6.71       \\
SGH      & \textbf{1.37}       & 1.60         & 2.46       &\textbf{ 6.41}      & 11.11       & 16.33     & 4.05       & 5.58         & \textbf{3.20}       \\
NPH      & \textbf{3.90}       & 5.62         & 8.62       & 5.37      & 4.00        & \textbf{3.97}      & 1.40       & \textbf{1.22}         & 2.15       \\
KCH      & 5.03       & 4.68         & \textbf{3.84}       & \textbf{1.74}      & 3.59        & 7.64      & \textbf{2.31}       & 3.21         & 3.91       \\
RLH      & \textbf{3.24}       & 4.88         & 7.43       & \textbf{2.59}      & 5.29        & 8.86      & 1.39       & \textbf{1.06}         & 1.36       \\
National & \textbf{14.35}      & 43.51        & 63.25      & \textbf{46.47}     & 120.59      & 324.59    & \textbf{25.19}      & 39.35        & 123.57     \\ \bottomrule
\end{tabular}
\end{table}

\subsection{Online monitoring}
It is vital to continuously monitor the performance after a machine-learning system is deployed. 
This is to prevent two possible scenarios (1) the gradual change in feature distribution worsens the predictive performance and (2) the breaking change in the upstream data pipelines causes data quality issues.
Therefore, we always validate the model using a held-out set and record the performance whenever a model is re-trained.
We have automated this process as part of the training procedure and developed a dashboard to visually track the performance over time.

\section{Illustrative use case}
\label{sec:usecase}

\begin{figure}
  \includegraphics[width=1\columnwidth]{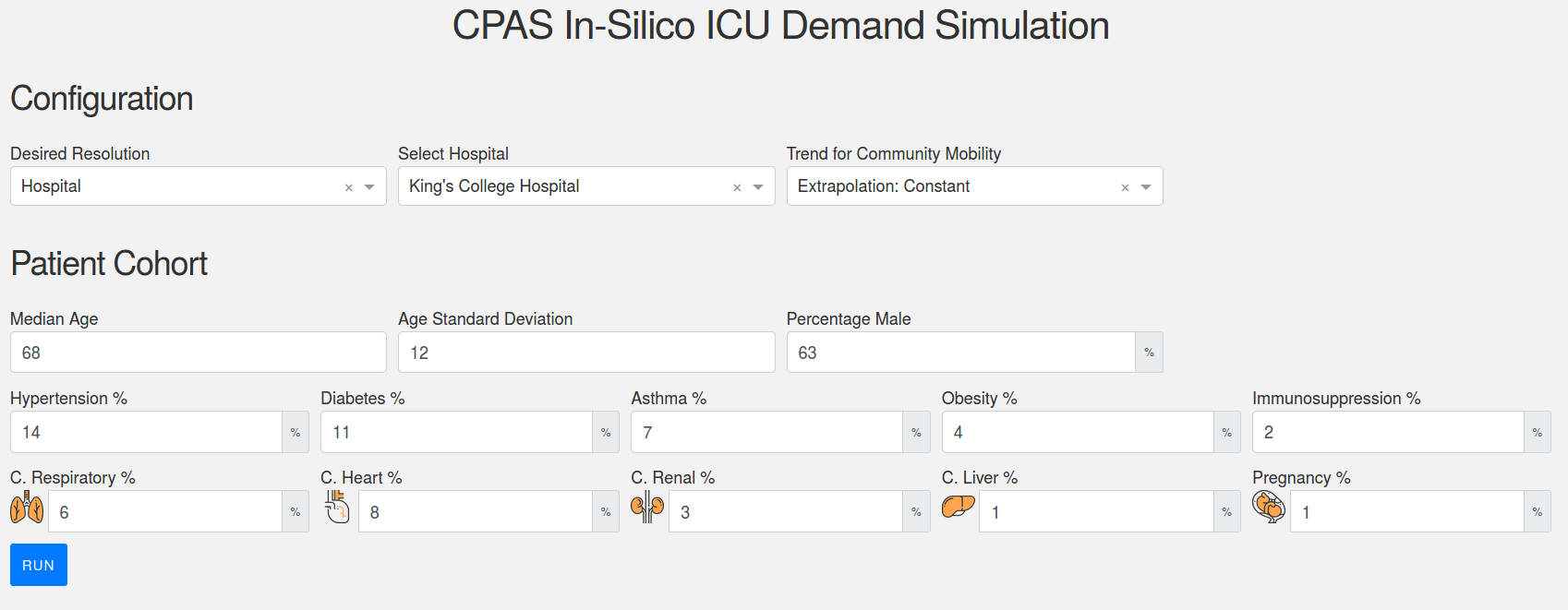}
\caption{The configuration interface of CPAS. The user enters the desired level of resolution and the region of interest. The user then inputs the assumed trend for future community mobility. The empirical feature distribution in the region of interest is displayed below for reference. }
\label{fig:6}       % Give a unique label
\end{figure}

\begin{figure}
  \includegraphics[width=1\columnwidth]{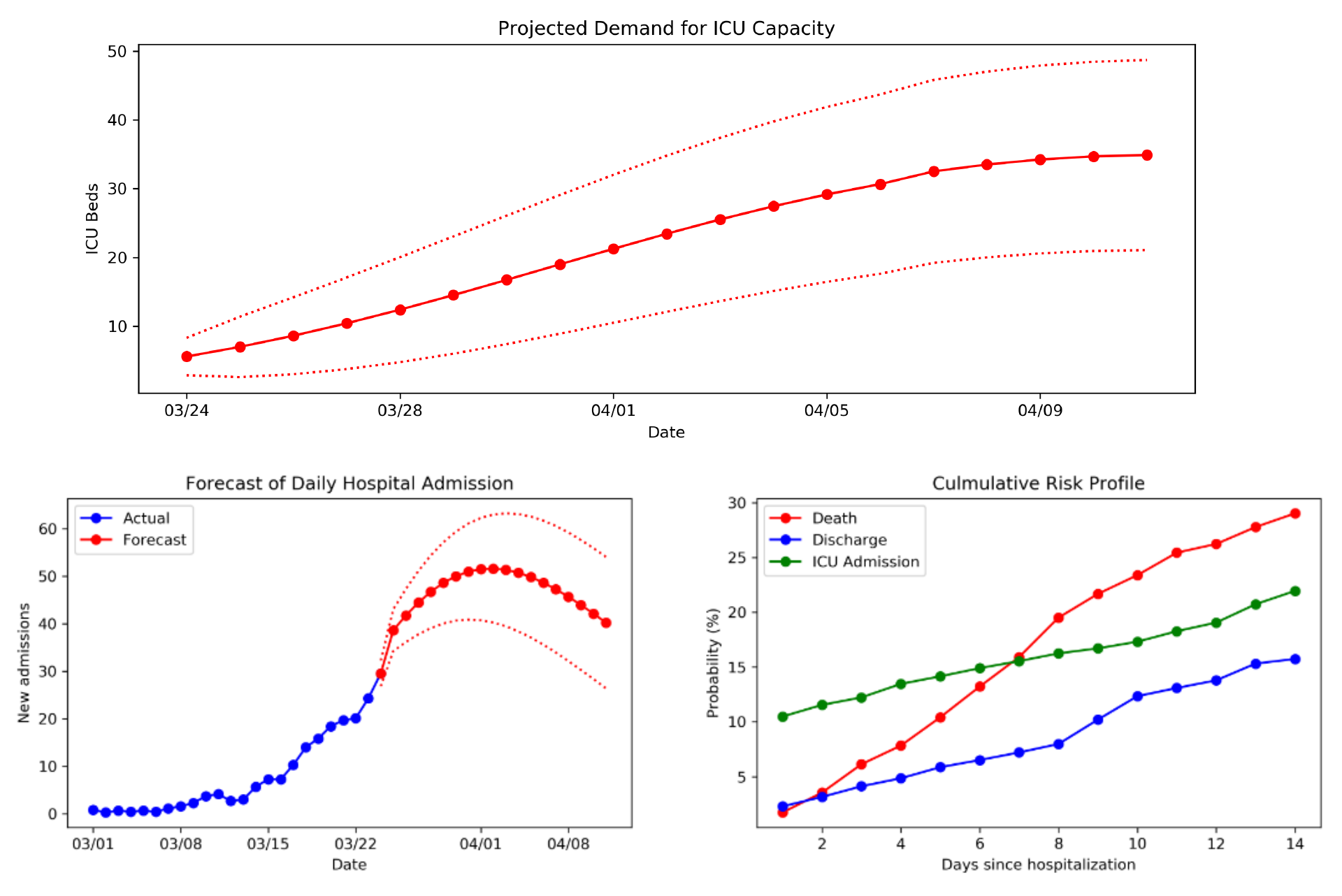}
\caption{The output interface of CPAS. CPAS displays the projected ICU demand with confidence intervals on the top. It then shows the intermediate prediction that leads to the projections. On the left, it shows the output of the aggregated trend forecaster. On the right, it shows the average risk profile for various outcomes given by the individualized risk predictor.}
\label{fig:7}       % Give a unique label
\end{figure}

Here we present an illustrative use case to demonstrate how CPAS works in real life.
In this example, we show how the management of a particular hospital located in central London can use CPAS to plan ICU surge capacity for future patients before the peak of COVID-19. 
On March 23\textsuperscript{th}, the hospital's ICU capacity had almost been fully utilized, but the pandemic had not yet reached the peak (refer to Figure \ref{fig:4}).
The hospital management was desperate to know how many more patients would be admitted to the hospital and to the ICU in the coming weeks.
They were planning to convert some of the existing general wards into ICU wards and, if necessary, to send some of the patients to the Nightingale hospital, a hospital specially constructed to support all NHS London hospitals in the surge of COVID-19 \citep{nightingale}. CPAS could help the management to estimate how many general wards to convert and how many patients to transfer.

% \paragraph{User input.}
Figure \ref{fig:6} presents the input interface of CPAS\footnote{All figures presented in this section are based on developmental versions of CPAS. They are for illustrative purposes only and do not represent the actual ``look-and-feel'' of CPAS in production.}.
The user first informed the system that the simulation should be performed on \textit{hospital} level. 
Since the peak of the pandemic had not occurred yet on March 23\textsuperscript{th}, it is reasonable to assume that the lockdown would continue and the community mobility would maintain at the post-lockdown level.
Therefore, the user selected the \textit{Extrapolation - Constant} option for the community mobility trend. 
Alternatively, the user can specify their own community mobility projections by selecting \textit{User-defined} in the drop down and upload a file with the forecast numbers.
CPAS displays the estimated feature distribution in the region of interest in the ``Patient Cohort'' section for reference.

The result of the simulation is shown in Figure \ref{fig:7}.
The panel on the top projects that the ICU demand in the hospital would increase over the next two weeks as the pandemic progresses. 
The ICU capacity for thirty five additional patients will be needed by mid April with the best scenario of 20 patients and the worst scenario of 50 patients.
In reality, the hospital admitted 42 patients to the ICU in this period, which falls within the confidence region given by CPAS.
With this estimation, the management decided that the hospital could cope with the surge by converting existing general wards into ICU wards, and there was probably no need to transfer patients to the Nightingale hospital. This also turned out to be the case in reality.

The ICU demand projection above is driven by the aggregated hospital admission forecast (Figure \ref{fig:7} bottom left), which predicts a steady increase of hospital admissions (around 50 per day) until early April when the trend starts to decline. The decline in admission is caused by the social distancing policy and the lowered community mobility, which are assumed to persist throughout the simulation. The projection is also driven by the predictions of individual risk profiles. The plot on bottom right shows the average risk profile of the patient cohort. There is significant risk for ICU admission from the first day of hospitalization, whereas the risk for death and discharge increases over time starting from a small value. By presenting these intermediate results, CPAS shows more transparency in the overall ICU demand projection.

\section{Lessons learned}
\label{sec:6}

AI and machine learning have certainly not moved slowly in bringing seismic change to countless areas including retail, logistics, advertising, and software development. But in healthcare, there is still great unexploited potential for systematic change and fundamental innovation. As in the CPAS project, we can use AI and machine learning to empower medical professionals by enhancing the guidance and information available to them.

Collaboration is one of the most important aspects of straddling the divide between machine learning research and healthcare applications. In the CPAS project, we work closely with clinicians and stakeholders because they brings in domain expertise to inform the formulation of the problem and the design of the system. Effective collaboration is a challenge as we are all highly specialized in our respective areas, with different ways of thinking and different professional languages and approaches. As a result, we must each make extra effort to reach the middle ground. 
But it’s a fascinating and invigorating way to work. Listening to clinicians and stakeholders can guide us to where problems and challenges actually lie, and then we can start being creative in trying to solve them.
We found it is particularly helpful to build prototypes rapidly and get timely feedback from the collaborators.
Each prototype should clearly demonstrate what can be achieved and what assumptions are made.
We can then iteratively find out what functionalities we should focus on most and what we can assume about the problem. 

Linking and accessing data is another challenge in healthcare applications. 
In the CPAS project, we first surveyed the potential data sources and understand the associated cost, which includes the financial cost, the waiting time for approval, the engineering cost, and so on.
After a thorough discussion with the collaborators, we then made a clear road map and prioritize what data to acquire first.
It is often practical to start out with a single easy-to-access data source, and then expand the data sources as the system gets more adoption. 
The success of CPAS is greatly facilitated by the solid data infrastructure in the UK's healthcare system. 
Initiatives such as CHESS play a vital role in the data-driven response to the pandemic. 

Transparency and interpretability is crucial for high-stake machine learning applications.
The reality is that most machine learning models can’t be used as-is by medical professionals because, on their own, they are black boxes that are hard for the intended users to apply, understand, and trust.
While interpretable machine learning \citep{ahmad2018interpretable} is still an open research area, CPAS explored two practical approaches to make a machine learning system more interpretable. 
The first approach is to break down the problem into a set of sub-problems. This divide-and-conquer approach allows the users to understand how the final answer is derived from the smaller problems. 
The second approach is to let the users autonomously explore different scenarios using simulation rather than presenting the result as the only possible answer. Scenario analysis  also allows the uses to understand the level of uncertainty and sensitivity of the machine learning predictions.

Last but equally importantly, automated machine learning is a powerful tool for building large and complex machine learning systems. Most machine learning models cannot be easily used off-the-shelf with the default hyperparameters. 
Moreover, there are many machine learning algorithms to choose from, and selecting which one is best in a particular setting is non-trivial – the results depend on the characteristics of the data, including number of samples, interactions among features and among features and outcomes, as well as performance metrics used. In addition, in any practical application, we need entire processing pipelines which involve imputation, feature selection, prediction, and calibration. AutoML is essential in order to enable machine learning to be applied effectively and at scale given the complexities stated above. In CPAS, we used AutoML to generate a large number of machine learning models with minimum manual tweaking. AutoML not only helps the CPAS models to issue more accurate predictions but also saves the developers manual work so that we can focus on the design and modelling aspects. 
In the future, we will also explore the usage of AutoML to address the temporal shift in the feature distribution \citep{zhang2019lifelong}, and to model time series collected in the clinical setting \citep{time}.

% the system designer should be well-versed in various sub-fields of machine learning and have a repertoire of learning algorithms. This will allows the practical problem to be properly formulated into a learning problem. In addition, when implementing systems that support high-stake decisions, it is preferred to use well-proven algorithms rather than experimental ones due to robustness considerations.

\begin{acknowledgements}
We thank Professor Jonathan Benger, Chief Medical Officer of NHS Digital, Dr. Jem Rashbass, Executive Director of Master Registries and Data at NHS Digital, Emma Summers, Programme Delivery Lead at NHS Digital, and the colleagues from NHS Digital and Public Health England for their tremendous support in providing datasets, computation environment, and domain expertise.
\end{acknowledgements}

\section*{Conflict of interest}

The authors declare that they have no conflict of interest.

\bibliographystyle{spbasic}      % basic style, author-year citations
\bibliography{ref}   % name your BibTeX data base

\begin{thebibliography}{43}
\providecommand{\natexlab}[1]{#1}
\providecommand{\url}[1]{{#1}}
\providecommand{\urlprefix}{URL }
\expandafter\ifx\csname urlstyle\endcsname\relax
  \providecommand{\doi}[1]{DOI~\discretionary{}{}{}#1}\else
  \providecommand{\doi}{DOI~\discretionary{}{}{}\begingroup
  \urlstyle{rm}\Url}\fi
\providecommand{\eprint}[2][]{\url{#2}}

\bibitem[{Ahmad et~al.(2018)Ahmad, Eckert, and
  Teredesai}]{ahmad2018interpretable}
Ahmad MA, Eckert C, Teredesai A (2018) Interpretable machine learning in
  healthcare. In: Proceedings of the 2018 ACM international conference on
  bioinformatics, computational biology, and health informatics, pp 559--560

\bibitem[{Alaa and Schaar(2018)}]{alaa2018autoprognosis}
Alaa A, Schaar M (2018) Autoprognosis: Automated clinical prognostic modeling
  via bayesian optimization with structured kernel learning. In: International
  Conference on Machine Learning, pp 139--148

\bibitem[{Alaa and van~der Schaar(2018)}]{alaa2018prognostication}
Alaa AM, van~der Schaar M (2018) Prognostication and risk factors for cystic
  fibrosis via automated machine learning. Scientific reports 8(1):1--19

\bibitem[{Alaa et~al.(2019)Alaa, Bolton, Di~Angelantonio, Rudd, and van~der
  Schaar}]{alaa2019cardiovascular}
Alaa AM, Bolton T, Di~Angelantonio E, Rudd JH, van~der Schaar M (2019)
  Cardiovascular disease risk prediction using automated machine learning: A
  prospective study of 423,604 uk biobank participants. PloS one 14(5):e0213653

\bibitem[{Bedford et~al.(2020)Bedford, Enria, Giesecke, Heymann, Ihekweazu,
  Kobinger, Lane, Memish, Oh, Schuchat et~al.}]{bedford2020covid}
Bedford J, Enria D, Giesecke J, Heymann DL, Ihekweazu C, Kobinger G, Lane HC,
  Memish Z, Oh Md, Schuchat A, et~al. (2020) Covid-19: towards controlling of a
  pandemic. The Lancet 395(10229):1015--1018

\bibitem[{Buuren and Groothuis-Oudshoorn(2010)}]{buuren2010mice}
Buuren Sv, Groothuis-Oudshoorn K (2010) mice: Multivariate imputation by
  chained equations in r. Journal of statistical software pp 1--68

\bibitem[{Charlson et~al.(1994)Charlson, Szatrowski, Peterson, and
  Gold}]{charlson1994validation}
Charlson M, Szatrowski TP, Peterson J, Gold J (1994) Validation of a combined
  comorbidity index. Journal of clinical epidemiology 47(11):1245--1251

\bibitem[{Chen and Guestrin(2016)}]{chen2016xgboost}
Chen T, Guestrin C (2016) Xgboost: A scalable tree boosting system. In:
  Proceedings of the 22nd acm sigkdd international conference on knowledge
  discovery and data mining, pp 785--794

\bibitem[{Chernick et~al.(2011)Chernick, Gonz{\'a}lez-Manteiga, Crujeiras, and
  Barrios}]{chernick2011bootstrap}
Chernick MR, Gonz{\'a}lez-Manteiga W, Crujeiras RM, Barrios EB (2011) Bootstrap
  methods

\bibitem[{Cox(1972)}]{cox1972regression}
Cox DR (1972) Regression models and life-tables. Journal of the Royal
  Statistical Society: Series B (Methodological) 34(2):187--202

\bibitem[{De~Leeuw(1977)}]{de1977correctness}
De~Leeuw J (1977) Correctness of kruskal's algorithms for monotone regression
  with ties. Psychometrika 42(1):141--144

\bibitem[{Google(2020)}]{Mobility}
Google (2020) Covid-19 community mobility.
  \url{https://www.google.com/covid19/mobility/}, accessed: 2020-07-04

\bibitem[{Guyon et~al.(2002)Guyon, Weston, Barnhill, and
  Vapnik}]{guyon2002gene}
Guyon I, Weston J, Barnhill S, Vapnik V (2002) Gene selection for cancer
  classification using support vector machines. Machine learning
  46(1-3):389--422

\bibitem[{Hethcote(2000)}]{hethcote2000mathematics}
Hethcote HW (2000) The mathematics of infectious diseases. SIAM review
  42(4):599--653

\bibitem[{Hinton(1990)}]{hinton1990connectionist}
Hinton GE (1990) Connectionist learning procedures. In: Machine learning,
  Elsevier, pp 555--610

\bibitem[{Hothorn et~al.(2006)Hothorn, B{\"u}hlmann, Dudoit, Molinaro, and Van
  Der~Laan}]{hothorn2006survival}
Hothorn T, B{\"u}hlmann P, Dudoit S, Molinaro A, Van Der~Laan MJ (2006)
  Survival ensembles. Biostatistics 7(3):355--373

\bibitem[{Hutter et~al.(2019)Hutter, Kotthoff, and
  Vanschoren}]{hutter2019automated}
Hutter F, Kotthoff L, Vanschoren J (2019) Automated machine learning: methods,
  systems, challenges. Springer Nature

\bibitem[{Hutzenthaler et~al.(2011)Hutzenthaler, Jentzen, and Kloeden}]{EULER}
Hutzenthaler M, Jentzen A, Kloeden PE (2011) Strong and weak divergence in
  finite time of euler's method for stochastic differential equations with
  non-globally lipschitz continuous coefficients. Proceedings of the Royal
  Society A: Mathematical, Physical and Engineering Sciences
  467(2130):1563--1576

\bibitem[{Hyvarinen(1999)}]{hyvarinen1999fast}
Hyvarinen A (1999) Fast ica for noisy data using gaussian moments. In: 1999
  IEEE International Symposium on Circuits and Systems (ISCAS), IEEE, vol~5, pp
  57--61

\bibitem[{Kermack and McKendrick(1927)}]{kermack1927contribution}
Kermack WO, McKendrick AG (1927) A contribution to the mathematical theory of
  epidemics. Proceedings of the royal society of london Series A, Containing
  papers of a mathematical and physical character 115(772):700--721

\bibitem[{Kingma and Ba(2014)}]{ADAM}
Kingma DP, Ba J (2014) Adam: A method for stochastic optimization. arXiv
  preprint arXiv:14126980

\bibitem[{Kotthoff et~al.(2017)Kotthoff, Thornton, Hoos, Hutter, and
  Leyton-Brown}]{kotthoff2017auto}
Kotthoff L, Thornton C, Hoos HH, Hutter F, Leyton-Brown K (2017) Auto-weka 2.0:
  Automatic model selection and hyperparameter optimization in weka. The
  Journal of Machine Learning Research 18(1):826--830

\bibitem[{Lee et~al.(2019)Lee, Zame, Alaa, and Schaar}]{lee2019temporal}
Lee C, Zame W, Alaa A, Schaar M (2019) Temporal quilting for survival analysis.
  In: The 22nd International Conference on Artificial Intelligence and
  Statistics, pp 596--605

\bibitem[{Li and Muldowney(1995)}]{li1995global}
Li MY, Muldowney JS (1995) Global stability for the seir model in epidemiology.
  Mathematical biosciences 125(2):155--164

\bibitem[{Liaw et~al.(2002)Liaw, Wiener et~al.}]{liaw2002classification}
Liaw A, Wiener M, et~al. (2002) Classification and regression by randomforest.
  R news 2(3):18--22

\bibitem[{NHS(2020{\natexlab{a}})}]{ICM}
NHS (2020{\natexlab{a}}) Health careers in intensive care medicine.
  \url{https://www.healthcareers.nhs.uk/explore-roles/doctors/roles-doctors/intensive-care-medicine
  }, accessed: 2020-07-04

\bibitem[{NHS(2020{\natexlab{b}})}]{ICU}
NHS (2020{\natexlab{b}}) Intensive care.
  \url{https://www.nhs.uk/conditions/Intensive-care/}, accessed: 2020-07-04

\bibitem[{NHS(2020{\natexlab{c}})}]{nightingale}
NHS (2020{\natexlab{c}}) Nhs nightingale london hospital.
  \url{http://www.bartshealth.nhs.uk/nightingale}, accessed: 2020-07-04

\bibitem[{NHS(2020{\natexlab{d}})}]{CPAS}
NHS (2020{\natexlab{d}}) Trials begin of machine learning system to help
  hospitals plan and manage covid-19 treatment resources developed by nhs
  digital and university of cambridge.
  \url{https://digital.nhs.uk/news-and-events/news/trials-begin-of-machine-learning-system-to-help-hospitals-plan-and-manage-covid-19-treatment-resources-developed-by-nhs-digital-and-university-of-cambridge},
  accessed: 2020-06-28

\bibitem[{Osemwinyen and Diakhaby(2015)}]{osemwinyen2015mathematical}
Osemwinyen AC, Diakhaby A (2015) Mathematical modelling of the transmission
  dynamics of ebola virus. Applied and Computational Mathematics 4(4):313--320

\bibitem[{Platt et~al.(1999)}]{platt1999probabilistic}
Platt J, et~al. (1999) Probabilistic outputs for support vector machines and
  comparisons to regularized likelihood methods. Advances in large margin
  classifiers 10(3):61--74

\bibitem[{P{\"o}lsterl et~al.(2016)P{\"o}lsterl, Navab, and
  Katouzian}]{polsterl2016efficient}
P{\"o}lsterl S, Navab N, Katouzian A (2016) An efficient training algorithm for
  kernel survival support vector machines. arXiv preprint arXiv:161107054

\bibitem[{Railsback et~al.(2006)Railsback, Lytinen, and
  Jackson}]{railsback2006agent}
Railsback SF, Lytinen SL, Jackson SK (2006) Agent-based simulation platforms:
  Review and development recommendations. Simulation 82(9):609--623

\bibitem[{Ranganath et~al.(2014)Ranganath, Gerrish, and
  Blei}]{ranganath2014black}
Ranganath R, Gerrish S, Blei D (2014) Black box variational inference. In:
  Artificial Intelligence and Statistics, pp 814--822

\bibitem[{Rasmussen(2003)}]{rasmussen2003gaussian}
Rasmussen CE (2003) Gaussian processes in machine learning. In: Summer School
  on Machine Learning, Springer, pp 63--71

\bibitem[{van~der Schaar et~al.(2020)van~der Schaar, Yoon, Qian, Jarrett, and
  Bica}]{time}
van~der Schaar M, Yoon J, Qian Z, Jarrett D, Bica I (2020) clairvoyance alpha:
  the first unified end-to-end automl pipeline for time-series data.
  \url{https://www.vanderschaar-lab.com/clairvoyance-alpha-the-first-unified-end-to-end-automl-pipeline-for-time-series-data/},
  accessed: 2020-07-04

\bibitem[{Snoek et~al.(2012)Snoek, Larochelle, and Adams}]{snoek2012practical}
Snoek J, Larochelle H, Adams RP (2012) Practical bayesian optimization of
  machine learning algorithms. In: Advances in neural information processing
  systems, pp 2951--2959

\bibitem[{Stekhoven and B{\"u}hlmann(2012)}]{stekhoven2012missforest}
Stekhoven DJ, B{\"u}hlmann P (2012) Missforest—non-parametric missing value
  imputation for mixed-type data. Bioinformatics 28(1):112--118

\bibitem[{Van~Belle et~al.(2011)Van~Belle, Pelckmans, Suykens, and
  Van~Huffel}]{van2011learning}
Van~Belle V, Pelckmans K, Suykens JA, Van~Huffel S (2011) Learning
  transformation models for ranking and survival analysis. Journal of machine
  learning research 12(3)

\bibitem[{Wingate and Weber(2013)}]{wingate2013automated}
Wingate D, Weber T (2013) Automated variational inference in probabilistic
  programming. arXiv preprint arXiv:13011299

\bibitem[{Yoon et~al.(2018)Yoon, Jordon, and Van Der~Schaar}]{yoon2018gain}
Yoon J, Jordon J, Van Der~Schaar M (2018) Gain: Missing data imputation using
  generative adversarial nets. International Conference on Machine Learning
  (ICML)

\bibitem[{Zhang et~al.(2019)Zhang, Jordon, Alaa, and van~der
  Schaar}]{zhang2019lifelong}
Zhang Y, Jordon J, Alaa AM, van~der Schaar M (2019) Lifelong bayesian
  optimization. arXiv preprint arXiv:190512280

\bibitem[{Zou and Hastie(2005)}]{zou2005regularization}
Zou H, Hastie T (2005) Regularization and variable selection via the elastic
  net. Journal of the royal statistical society: series B (statistical
  methodology) 67(2):301--320

\end{thebibliography}

% \begin{thebibliography}{}
% %
% % and use \bibitem to create references. Consult the Instructions
% % for authors for reference list style.
% % Format for Journal Reference
% Author, Article title, Journal, Volume, page numbers (year)
% Author, Book title, page numbers. Publisher, place (year)
% \end{thebibliography}

\end{document}